\documentclass{article}

    \usepackage[final]{neurips_2020_ml4ad}

\usepackage[utf8]{inputenc} 
\usepackage[T1]{fontenc}    
\usepackage{hyperref}       
\usepackage{url}            
\usepackage{booktabs}       
\usepackage{amsfonts}       
\usepackage{nicefrac}       
\usepackage{microtype}      
\usepackage{algorithm} 
\usepackage{algpseudocode}
\usepackage{amsmath}
\usepackage{amssymb}
\usepackage{graphicx}
\usepackage{caption, subcaption}
\usepackage[skip=3ex]{caption}
\usepackage{float}
\usepackage{xcolor}

\title{Certified Interpretability Robustness for Class Activation Mapping}

\newcommand{\R}{\mathbb{R}}

\author{%
  Alex Gu \\
  MIT \\ Cambridge, MA \\ \texttt{gua@mit.edu}
  \And
  Tsui-Wei Weng \\ MIT-IBM Watson AI Lab \\ IBM Research \\ \texttt{twweng@mit.edu}
  \And
  Pin-Yu Chen \\
  MIT-IBM Watson AI Lab \\ IBM Research \\ \texttt{pin-yu.chen@ibm.com}
  \And
  Sijia Liu \\
  MIT-IBM Watson AI Lab \\ IBM Research \\ \texttt{sijia.liu@ibm.com}
  \And
  Luca Daniel \\  MIT \\ Cambridge, MA \\ \texttt{luca@mit.edu}
}

\begin{document}

\maketitle

\begin{abstract}
  Interpreting machine learning models is challenging but crucial for ensuring the safety of deep networks in autonomous driving systems. Due to the prevalence of deep learning based perception models in autonomous vehicles, accurately interpreting their predictions is crucial. While a variety of such methods have been proposed, most are shown to lack robustness. Yet, little has been done to provide certificates for interpretability robustness. Taking a step in this direction, we present CORGI, short for \textbf{C}ertifiably pr\textbf{O}vable \textbf{R}obustness \textbf{G}uarantees for \textbf{I}nterpretability mapping. CORGI is an algorithm that takes in an input image and gives a certifiable lower bound for the robustness of the top $k$ pixels of its CAM interpretability map. We show the effectiveness of CORGI via a case study on traffic sign data, certifying lower bounds on the minimum adversarial perturbation not far from (4-5x) state-of-the-art attack methods.
\end{abstract}

\section{Introduction}

Deep neural networks are commonplace in safety-critical instances such as real-world autonomous driving systems. However, a major issue surrounding these networks is that they lack robustness to adversarial perturbations~\citep{szegedy2013intriguing}. Under visually imperceptible perturbations, well-trained models can easily be fooled into confidently making predictions for an incorrect class, even in the black-box setting when the adversary does not even have access to the network architecture~\citep{guo2019simple}. In the past few years, we have seen a series of increasingly stronger attacks against neural networks (Fast Gradient Sign Method~\citep{goodfellow2014explaining}, Projected Gradient Descent~\citep{kurakin2016adversarial}, Carlini-Wagner~\citep{carlini2017towards}). 

A dual view of robustness research has focused on providing provable defenses against these attacks. For a neural network $NN$ and some input image $x$, attacks aim to find an adversarial example $x'$ such that $NN(x) \ne NN(x')$ and $\|x-x'\|_{\infty}=\epsilon$ is as small as possible. In contrast, the goal of these provable defenses is to provide a \textit{certifiable radius} $r$ that would guarantee that $NN(x) = NN(x')$ for all $\|x-x'\|_{\infty} \le r$. While being able to determine the maximum certificate $r^*$ would provide us optimal levels of safety, doing so, even under an approximation ratio, has been shown to be hard~\citep{weng2018towards}. However, various approaches have been used to come close to this goal, such as semidefinite programming~\citep{raghunathan2018semidefinite}, mixed-integer programming~\citep{tjeng2017evaluating}, and linear bounding of activation functions~\citep{boopathy2019cnn}.

A frightening concern is that while deep neural networks make incorrect predictions on adversarial examples, they leave little to no insight towards what went wrong. As autonomous vehicle systems are comprised of deep neural networks for perception systems~\citep{chen2020end}, being able to interpret their predictions is crucial in establishing trust and understanding between humans and these systems. This issue has given rise to many interpretability methods including CAM ~\citep{zhou2016learning}, GradCAM~\citep{selvaraju2017grad}, Integrated Gradients~\citep{sundararajan2017axiomatic}, SmoothGrad~\citep{smilkov2017smoothgrad}, and DeepLIFT~\citep{shrikumar2017learning}. All of these methods generate an interpretability map for each image showing the effect of each input pixel on the overall prediction.

Unfortunately, recent work has shown that these interpretability methods are also prone to adversarial perturbations~\citep{zheng2019analyzing}~\citep{alvarez2018robustness}~\citep{ghorbani2019interpretation}. All three works successfully generate visually imperceptible perturbations for which the network correctly classifies the image, but shows a nonsensical interpretability map. While providing certifiable guarantees for adversarial attacks on a neural network classifiers has been thoroughly studied~\citep{singh2019abstract},~\citep{zhang2018efficient}, little has been done to certify robustness for interpretability methods. 

The most relevant work to ours is~\citep{levine2019certifiably}, which certifies interpretability on a modification of SmoothGrad. While we follow their example and use the top-$k$ metric for interpretability, our work focuses on CAM, a well-established visual intepretability algorithm that not only provides input attribution importance, but also visually localizes class-discriminative regions. CAM is better suited for autonomous vehicle applications, where being able to cleanly distinguish similar road signs is essential for accident prevention. For example, as shown in Figure 1, the CAM map shows that the neural network is seeing the white stripe in the no entry sign, which is unique to that class. In addition, while computation of SmoothGrad requires a backward pass for each test image, CAM only requires a forward pass. Given the importance of interpretability methods like CAM, we propose an algorithm for the interpretability robustness of CAM, as being able to guarantee a certain degree of safety is imperative for our trust in autonomous driving systems.

Our paper contains the following contributions:
\begin{itemize} 
\item We present CORGI (\textbf{C}ertifiably pr\textbf{O}vable \textbf{R}obustness \textbf{G}uarantees for \textbf{I}nterpretability mapping), the first algorithm giving a lower bound on the interpretability robustness for CAM.

\item We provide two verifiable theoretical guarantees with corresponding experiments for an exact top-$k$ metric and relaxed top-$k$ metric, as well as a third theoretical result in terms of the Lipschitz constant of the neural network.

\item We empirically compare our CORGI lower bounds to Ghorbani's attack method's upper bounds~\citep{ghorbani2019interpretation} on the GTSRB traffic~\citep{stallkamp2012man} dataset, showing that on average, our CORGI lower bounds are only a factor of 4-5x away from current attack upper bounds.

\end{itemize}

\begin{figure}
\centering
\includegraphics[width=\linewidth]{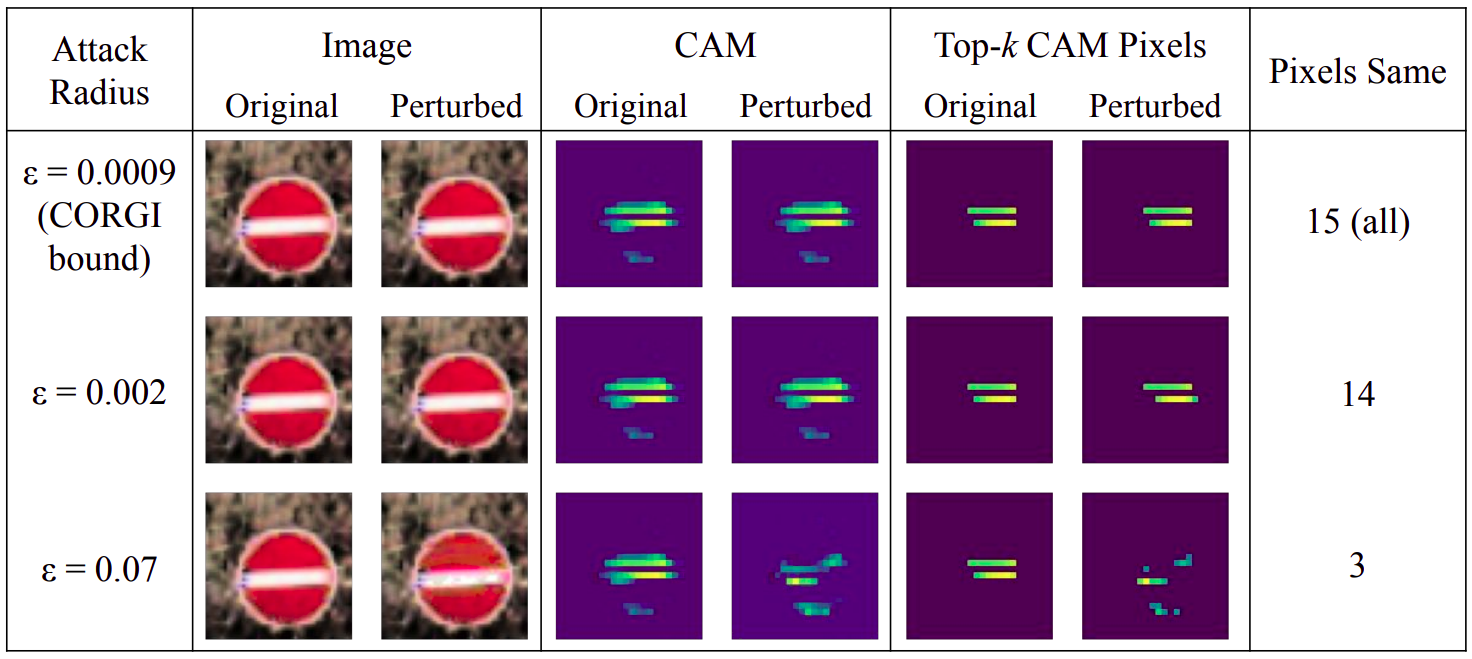}
\caption{An image in the GTSRB dataset under 3 different attack radii. Shown are the original and perturbed images, CAM maps, and the top-$k$ CAM pixel positions ($k=15$). Under our CORGI certificate, all top-$k$ pixels remain identical.}
\end{figure}
\section{Background and Notation}

In this section, we introduce our notation, explain how CAM interpretability maps are generated, and describe the attack algorithm~\citep{ghorbani2019interpretation} we use as an complementary upper bound.

\subsection{Notation}
For a neural network, input image $x \in \R^{d \times d}$, and arbitrary class $c$, let $I^c(x) \in \R^{s \times s}$ be the \textit{interpretability map} for input image $x$ and class $c$ generated by CAM (described in Section 2.2). In this work, we are interested in the CAM map only for the predicted class, so we will often drop the $c$ and write $I(x)$, understood to mean the interpretability map of input image $x$ for its predicted class. 

In addition, let $I(x)_1, I(x)_2, \cdots, I(x)_{s^2} \in \R$ be the values of the individual pixels of $I(x)$, and $\sigma_1, \sigma_2, \cdots, \sigma_{s^2}$ be a permutation of $1, 2, \cdots, s^2$ such that $I(x)_{\sigma_1} \ge I(x)_{\sigma_2} \ge \cdots \ge I(x)_{\sigma_{s^2}}$. Therefore, the $\sigma_i$'s denotes indices of the pixels in sorted order. For ease of notation, we will use $I(x)_{\sigma_i}$ interchangeably with $I(x)_{[i]}$ to denote the value of the $i$th largest pixel of the interpretability map, with $\sigma_i$ being the position of that pixel in the original map.

\subsection{Class Activation Mapping}
One popular technique used to generate interpretability maps for deep neural networks is known as \textit{class activation mapping}~\citep{zhou2016learning}. A class activation map (CAM) is a 2-dimensional heatmap that explains a neural network's prediction on an image by highlighting the parts of that image contributing to its prediction. A major advantage of CAM is that they can visually localize class-discriminative regions, which helps in understanding how a network distinguishes between similar classes.

CAM only provides an interpretability map for convolutional neural network architectures with their last convolutional layer followed by a global average pooling layer and a fully connected layer. Formally, consider such a network with last convolutional layer output of size $s \times s \times f$, where $s \times s$ is the size of the map, and $f$ is the number of filters. Fix an input image, and let $f^k_{x, y} \in \mathbb{R}$ represent the value of spatial location $(x, y)$ of filter $k$ in the last convolutional layer. Let $F^k \in \mathbb{R}$ be the output of the $k$th filter after the global average pooling layer, $$F^k = \sum_{1 \le x, y \le s} f^k_{x, y}$$ Now, let $w \in \mathbb{R}^{k}$ be the weights of the fully connected layer connected to the predicted class, which we denote as $c$. Therefore, the input to the final softmax is $\sum_{k} w_k^c F^k$, where $w_k^c$ are weights that intuitively capture the importance of feature map $k$ for the predicted class $c$. As a notational note, we will sometimes write this as $w_k$ when it is clear that $c$ is the predicted class. The class activation map for class $c$, $M^c \in \mathbb{R}^{s \times s}$, is computed as a weighted average of these feature maps: $$M^c_{x, y} = \sum_{k} w_k^c f^k_{x, y}$$

\subsection{Iterative Feature Importance Attack}
One property that we would like interpretability maps to have is that the top-$k$ largest pixels still remain fairly large under small perturbations. Denoting by $\sigma_i$ the position of the $i$th largest pixel in $I(x)$ and $\sigma'_i$ likewise in $I(x+\epsilon)$, we would like $\{\sigma_1, \sigma_2, \cdots, \sigma_k\} = \{ \sigma'_1, \sigma'_2 ,\cdots, \sigma'_k \}$. In other words, we want the $k$ largest pixels of the original image to remain the $k$ largest pixels in the perturbed image. A good attack would then ensure the number of $\sigma_i$'s that appear in $\{ \sigma'_1, \sigma'_2 ,\cdots, \sigma'_k \}$ is small.

The iterative feature importance attack~\citep{ghorbani2019interpretation} intuitively attacks features maps under the top-$k$ metric by finding a perturbation minimizing the sum of the values of the original top-$k$ pixel positions in the new image. Given an input image $x$, their attack first considers $B = \{\sigma_1, \sigma_2, \cdots, \sigma_{k} \}$, the top-$k$ pixel locations in $I(x)$. They then run projected FGSM (fast-gradient sign method)~\citep{goodfellow2014explaining} to minimize the objective $\sum_{i \in B} I(y)_i$ or equivalently, to maximize $D(x, y) = -\sum_{i \in B} I(y)_i$. In ReLU networks, however, where $\nabla_y D(x, y_{i-1})$ is piecewise-constant, we first replace the ReLU with a softplus as the original paper does. Therefore, $I(\cdot)$ is actually the interpretability map of the softplus network rather than the original ReLU network.

However, we make a small modification of their algorithm: at the end, instead of returning the $y$ with the largest dissimilarity, we return the $y$ such that $I(x)$ and $I(y)$ have the least number of top-$k$ pixels in common, as this is what CORGI certifies for. The algorithm is as follows:

\begin{algorithm}
    \caption{Gradient-based top-$k$ interpretability map attack~\citep{ghorbani2019interpretation}}
    \begin{algorithmic}[1]
        \Procedure{Top-k Attack}{$x$, $\epsilon$, $I(\cdot)$, $k$}
        \State Define $B = \{\sigma_1, \sigma_2, \cdots, \sigma_{k} \}$ to be the top-$k$ pixel locations of $I(x)$
        \State Define $D(x, y) = -\sum_{i \in B} I(y)_i$ (dissimilarity metric between interpretability maps of $x, y$)
        \State Initialize $y_0 = x$
        \For {$i = 1, \cdots, P$}
            \State $y_{i} = y_{i-1} + \alpha \cdot \text{sign}(\nabla_y D(x, y_{i-1}))$
            \State If necessary, clip $y_i$ to satisfy $|y_i-x|_{\infty} \le \epsilon$
        \EndFor
        \State \Return $y \in \{y_0, \cdots, y_P\}$ such that $I(x)$ and $I(y)$ have the least number of top-$k$ pixels in common
    \EndProcedure
    \end{algorithmic}
\end{algorithm}

\section{CORGI: A Framework for Certifiable Interpretability Robustness}
In this section, we present CORGI, a framework which provides a lower bound $r$ for the minimum adversarial attack radius $r^*$. When the interpretability attack method described in Section 2.3 is successful and returns an adversarial example $\|x'-x\|_{\infty} = \epsilon$, $\epsilon$ is then an upper bound for $r^*$. In contrast, we give a lower bound certificate $r$ such that for perturbed images $x'$ with $\|x-x'\|_{\infty} \le r$, the set of top-$k$ pixel positions of the original CAM map does not change. In other words, if $\{ \sigma'_i \}$ denotes the top $k$ pixel positions of the CAM map $I(x')$, then $\{\sigma_1, \cdots, \sigma_k \} = \{ \sigma'_1, \cdots, \sigma'_k \}$. Note that the ordering of these pixel positions does not necessarily need to be the same, just that the unordered set stay constant. One subtlety to mention is that our CORGI bound does not guarantee that the classifier itself is robust. However, we can take the minimum of our CORGI bound and the robustness bound to give a certificate that classification and interpretation are both accurate. Empirically, however, robust interpretability seems to be a more challenging task than robust classification, and we observed that CORGI bounds are almost always smaller than robustness bounds.

CORGI builds off of the CNN-Cert framework~\citep{boopathy2019cnn}, which provides efficient layer-wise bounds on convolutional neural networks. At a high level, CORGI operates in 2 stages. First, it uses CNN-Cert to give lower and upper bounds $L(\delta), R(\delta) \in \R^{s \times s}$ on the CAM map: $$L(\delta)_{i, j} \triangleq \min_{\|x'-x\|_{\infty} \le \delta} I(x')_{i, j}, U(\delta)_{i, j} \triangleq \max_{\|x'-x\|_{\infty} \le \delta} I(x')_{i, j}$$
Now, observe that guaranteed top-$k$ interpretability robustness under perturbations of radius at most $r$ implies robustness under any smaller perturbation radius $r'\le r$. Therefore, in the second stage, CORGI uses binary search to find the largest radius $r$ that it is able to accurately certify, giving us a lower bound certificate on interpretability robustness.

\subsection{Upper and Lower Bounding CAM Values with CNN-Cert}
Given a neural network and an input, CNN-Cert is a framework that is able to lower and upper bound all of the layers of a neural network under a maximum perturbation radius $\delta$. We use CNN-Cert as a subroutine to lower and upper bound CAM maps. First, we lower and upper bound the last convolutional layer output. Let $l_k(\delta), u_k(\delta) \in \R^{s \times s}$ denote the bounds of the $k$th filter of the last convolutional layer output $f^k$, so that $l_k(\delta)_{i, j} \le f^k(x')_{i, j} \le u_k(\delta)_{i, j}$ for all pixels $(i, j)$ and perturbed images $x'$ with $\|x'-x\|_{\infty} \le \delta$. We then combine these lower/upper bounds with the known weights $w_k$ of the fully connected layer to bound the CAM. To see how to do this, observe that for all $\|x'-x\|_{\infty} \le \delta$, \begin{align*} I(x'
)_{i, j} &= \sum_{k} w_k f^k(x')_{i, j}\\&= \sum_{k, w_k < 0} w_k f^k(x')_{i, j} +\sum_{k, w_k > 0} w_k f^k(x')_{i, j} \\&\ge  \sum_{k, w_k < 0} w_k u_k(\delta)_{i, j} + \sum_{k, w_k > 0} w_k l_k(\delta)_{i, j} \triangleq L(\delta)_{i, j} \addtocounter{equation}{1}\tag{\theequation} \end{align*}

Similarly, we have \begin{equation}I(x'
)_{i, j}\le  \sum_{k, w_k < 0} w_k l_k(\delta)_{i, j} + \sum_{k, w_k > 0} w_k u_k(\delta)_{i, j} \triangleq U(\delta)_{i, j}\end{equation}
Therefore, with these definitions of $L(\delta), U(\delta) \in \R^{s \times s}$, we then have, for all $\|x'-x\|_{\infty} \le \delta$, we now have an upper and lower bound on each pixel of the CAM map: \begin{equation} L(\delta)_{i, j} \le I(x')_{i, j} \le U(\delta)_{i, j} \end{equation}

\subsection{Testing for Certifiability}
The next stage of CORGI aims to find the largest perturbation radius that can be certified with the bounds in Eq. 3. For a fixed input image $x$ and a perturbation size $\delta$, a perturbed image $x'$ with $\|x-x'\|_{\infty}\le \delta$ will have the same top-$k$ pixels if and only if \begin{equation} \min_{k' \in 1, 2, \cdots, k} I(x')_{\sigma_{k'}} \ge \max_{k' \in k+1, \cdots, s^2} I(x')_{\sigma_{k'}} \end{equation}
We will find a condition that ensures that Eq. 4 holds. First, observe that since $I(x')_{\sigma_{k}} \ge L(\delta)_{\sigma_{k}}$ for all $\sigma_{k}$, \begin{equation} \min_{k' \in 1, 2, \cdots, k} I(x')_{\sigma_{k'}} \ge \min_{k' \in 1, 2, \cdots, k} L(\delta)_{\sigma_{k'}} \end{equation} 
Similarly, for the upper bound, \begin{equation}\max_{k' \in k+1, \cdots, s^2} U(\delta)_{\sigma_{k'}} \ge \max_{k' \in k+1, \cdots, s^2} I(x')_{\sigma_{k'}}\end{equation}

Eq. 5 and 6 then give rise to our theorem for certified CAM interpretability:
\paragraph{Theorem 3.2 (CAM Interpretability Certificate)} \textit{Let $x \in \R^{s \times s}$ be a fixed input image, $\delta \in \R_{\ge 0}$ be a fixed maximum perturbation size, $k \in \mathbb{Z}_{\ge 1}$, and $L(\delta), R(\delta)$ be CAM map bounds as computed in Eq. 1. Then, for all $x'$ such that $\|x-x'\|_{\infty} \le \delta$, if \begin{equation}\min_{k' \in 1, 2, \cdots, k} L(\delta)_{\sigma_{k'}} \ge \max_{k' \in k+1, \cdots, s^2} U(\delta)_{\sigma_{k'}}\end{equation} then $\delta$ is a certified lower bound of $r^*$. In other words, the top-$k$ pixel positions of the CAM map $I(x')$ must be the same as those in the original CAM map $I(x)$. }

\paragraph{Proof} This theorem follows since Eq. 4 follows directly from Eq. 5, 6, and the hypothesis Eq. 7: \begin{equation} \min_{k' \in 1, 2, \cdots, k} I(x')_{\sigma_{k'}} \ge \min_{k' \in 1, 2, \cdots, k} L(\delta)_{\sigma_{k'}} \ge \max_{k' \in k+1, \cdots, s^2} U(\delta)_{\sigma_{k'}} \ge  \max_{k' \in k+1, \cdots, s^2} I(x')_{\sigma_{k'}} \end{equation}

As a remark, observe that for $\delta=0$, the inequality is obviously true because CNN-Cert will return the values of the network at the original input, so $L(0) = I(x) = U(0)$, and $I(x)_{[k]} = \min_{k' \in 1, 2, \cdots, k} L(0)_{\sigma_{k'}} \ge \max_{k' \in k+1, \cdots, s^2} U(0)_{\sigma_{k'}} = I(x)_{[k+1]}$ by definition.

Now, notice that since the lower and upper bounds of CNN-Cert are monotonic in $\delta$, if $\delta' \le \delta$, then $L(\delta)_{i, j} \le L(\delta')_{i, j}$ and $U(\delta')_{i, j} \le U(\delta)_{i, j}$, so that $L(\delta)_{\sigma_i} \le L(\delta')_{\sigma_i}$ and $U(\delta')_{\sigma_i} \le U(\delta)_{\sigma_i}$ for all $1 \le i \le s^2$. Therefore, if Eq. 7 is satisfied for some $\delta$, then it is also satisfied for all $\delta' \ge \delta$. We can then binary search for the smallest certifiable radius.

\begin{algorithm}
    \caption{CORGI}
    \begin{algorithmic}[1]
        \Procedure{CORGI}{$x$, $f(\cdot)$, $k$}
        \State Use $f$ to compute the interpretability map $I(x)$
        \State Let $\sigma_1, \sigma_2, \cdots, \sigma_{s^2}$ be the indices of $I(x)$'s pixels from largest to smallest
        \State Initialize $\delta_l \leftarrow 0, \delta_u \leftarrow 1$
        \For {$i = 1, \cdots, P$}
            \State $\delta_m \leftarrow (\delta_l + \delta_u) / 2$
            \State Run CNN-Cert$(x, \delta_m)$ to get $l_k(\delta_m), u_k(\delta_m)$, last convolutional output layer bounds 
            \State Compute $L(\delta_m), U(\delta_m)$ from $l_k(\delta_m), u_k(\delta_m)$, and $f$'s weights $w_k$ (Eq. 1, 2)

            \If {Eq. 7 holds (which means $\delta_m$ is a certificate)}
                \State $\delta_{u} \leftarrow \delta_m$
            \Else
                \State $\delta_{l} \leftarrow \delta_m$
            \EndIf
        \EndFor
        \State \Return $\delta_u$ (an lower bound on the certificate)
    \EndProcedure
    \end{algorithmic}
\end{algorithm}

\subsection{A More Relaxed Interpretability Guarantee}
While requiring that the top-$k$ pixel positions remain the same is a nice condition to have, sometimes it will be enough for the top-$k$ pixel positions to remain relatively large in the perturbed image without remaining in the top-$k$. If $I(x)_{[k]}$ is very close in value to $I(x)_{[k+1]}$, requiring that the original top-$k$ pixels $\{\sigma_1, \cdots, \sigma_k\}$ be the same as the perturbed top-$k$ pixels $\{\sigma'_1, \cdots, \sigma'_k\}$ may be a strong condition. In this case, even if a perturbation swapped the relative values of pixels $\sigma_k$ and $\sigma_{k+1}$ causing $I(x+\delta)_{\sigma_k} < I(x+\delta)_{\sigma_{k+1}}$, the interpretability map would still remain fairly accurate. Therefore, we relax the original condition and instead only require that $\{\sigma_1, \cdots, \sigma_k\} \subseteq \{\sigma'_1, \cdots, \sigma'_{k_2} \}$, for some $k_2 \ge k$. Intuitively, this says that the top-$k$ pixel positions of the original image must remain in the top-$k_2$ pixel positions of the perturbed image. Note that $k_2=k$ corresponds to our original requirement that the top-$k$ pixel positions remain the same under perturbation. 

The algorithm for this scenario remains very similar to that in Section 3.2, and the crux is once again checking whether or not we can certify top-$k$ interpretability robustness for a given $\delta$. As before, we first compute $L(\delta), U(\delta)$ as in Section 3.1. Now, define $\tau_1, \cdots, \tau_{s^2-k}$ to be an ordering of all pixel positions in $S = \{1, 2, \cdots, s^2\} \setminus \{ \sigma_1, \sigma_2, \dots, \sigma_k\}$ satisfying $U(\delta)_{\tau_1} \ge U(\delta)_{\tau_2} \ge \cdots \ge U(\delta)_{\tau_{s^2-k}}$, and for notational simplicity define $U(\delta)_{\{k\}} = U(\delta)_{\tau_k}$. $U(\delta)_{\{k\}}$ then denotes the $k$th largest upper bound of all the pixels that weren't originally in the top-$k$ of $I(x)$. Similarly, define $U(\delta)_{[k]}$ to be the $k$th largest pixel value in $U(\delta)$ (no longer restricted on $S$) and $I(x')_{[k]}$to be the $k$th largest pixel value in the interpretability map $I(x')$. We can then prove a theorem analogous to Theorem 3.2:

\paragraph{Theorem 3.3 (Relaxed CAM Interpretability Certificate)} \textit{Let $x \in \R^{s \times s}$ be a fixed input image, $\delta \in \R_{\ge 0}$ be a fixed maximum perturbation size, $k, k_2 \in \mathbb{Z}_{\ge 1}$ such that $k_2 \ge k$, and $L(\delta), R(\delta)$ be CAM map bounds as computed in Eq. 1. Then, for all $x'$ such that $\|x-x'\|_{\infty} \le \delta$, if \begin{equation}\min_{k' \in 1, 2, \cdots, k} L(\delta)_{\sigma_{k'}} \ge U(\delta)_{\{k_2-k+1\}}\end{equation} then under all perturbations of size at most $\delta$, the top-$k$ pixel positions of the original CAM map $I(x)$ are in the top $k_2$ pixel positions of the CAM map $I(x')$.}

The proof is very similar to that of Theorem 3.2 and is deferred to the appendix. Using this theorem, we can simply run the same binary search as described in the previous section. The only modification that must be made to the original algorithm is that we must now check Eq. 9 instead of Eq. 7 in line 8 of CORGI.

\section{Experiments}

\paragraph{Dataset and Model}{We use the German Traffic Sign Recognition Benchmark (GTSRB)~\citep{stallkamp2012man}, a large, lifelike database of over $50000$ $48 \times 48 \times 3$ traffic signs in $43$ classes. We train a small convolutional neural network of four convolutional layers with $8$, $8$, $16$, and $16$ filters, a global average pooling layer, and a fully connected layer (as required by CAM).}

\paragraph{Interpretability Attack}{First, we ran CORGI with $k=15$ on a sample "no entry" sign to get a certified radius of $r=0.0009$. We first implemented a the attack algorithm (Algorithm 1) with $P=300$ and $\alpha = \epsilon/10$. Figure 1 shows the attack algorithm's perturbations on both $r$ and two larger values of $\epsilon$. The perturbed images, CAM maps, and top-$k$ maps are shown. Note that within our CORGI bound, all the top-$k$ pixels remain the same.}

\paragraph{Comparing CORGI Bounds to Interpretability Attack Bounds} By running binary search on the interpretability attack algorithm, we can find the maximum $\epsilon$ that yields a successful top-$k$ attack, giving us an upper bound on the best certificate $r^*$. Since CORGI is currently the first framework that provides interpretability robustness certificates for CAM, we compare CORGI's certificates with complementary upper bounds generated by the iterative feature importance attack. To do so, we sampled 100 random images from each of two classes, class 17 ("no entry") and class 12 ("priority road"). Figure 2a shows two samples of each image. 

Figure 2b shows both the CORGI lower bound and attack upper bound for each of the 200 images. For ease of comparison, the images in each class were sorted by their CORGI lower bound. For each of the 200 images, we also calculated the ratio $\epsilon/r$, where $\epsilon$ is the minimum perturbation radius found by the attack, and $r$ is the maximum CORGI certifiable radius (Figure 2c). There were a few outliers in both classes, where the attack bound was abnormally large. The figure removes these outliers and only plots samples for which the gap was under 20. The resulting figure consists of 99/100 class 12 samples and 92/100 of the class 17 samples. the majority of the ratios were fairly small, with a median of 5.0 for class 17 and 4.7 for class 12. This signifies that for this network, CORGI bounds are only 4-5 times larger than attack bounds, and therefore from the optimal perturbation $r^*$. We show additional examples of CORGI vs attack bounds for 5 other classes in the appendix.

\begin{figure}[htbp]
  \begin{subfigure}[t]{0.16\textwidth}
    \centering
    \includegraphics[width=\linewidth]{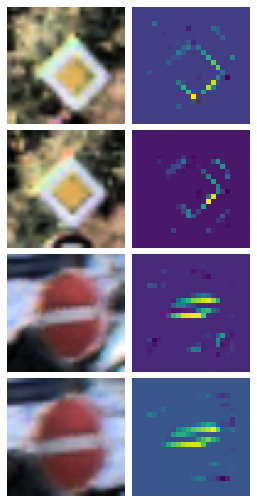}
    \caption{}
  \end{subfigure}
  \begin{subfigure}[t]{0.43\textwidth}
    \centering
    \includegraphics[width=\linewidth]{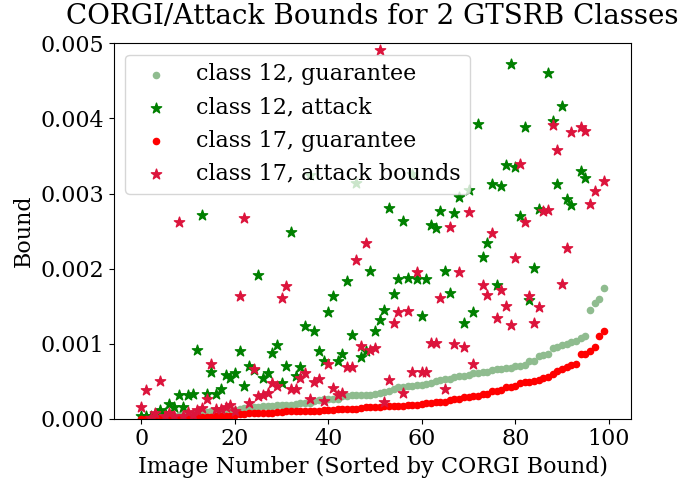}
    \caption{}
  \end{subfigure}
  \begin{subfigure}[t]{0.38\textwidth}
    \centering
    \includegraphics[width=\linewidth]{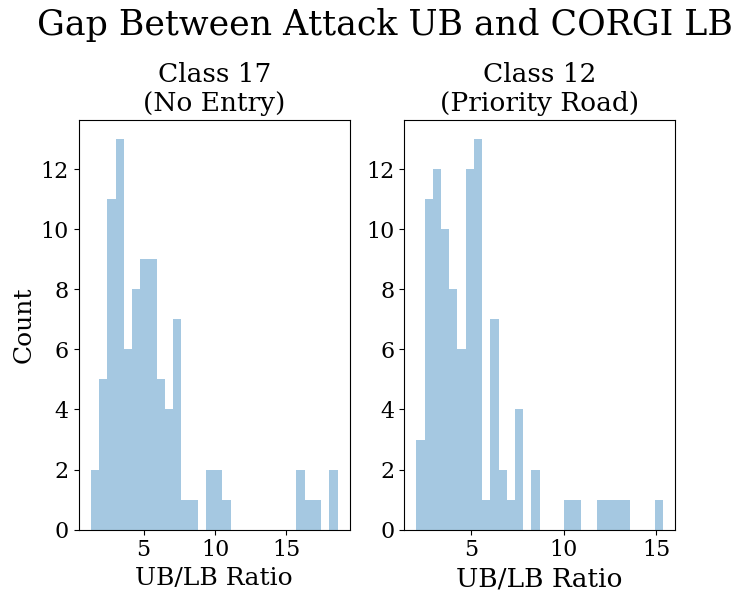}
    \caption{}
  \end{subfigure}
\centering
\caption{(a) Sample images and CAM maps from class 12 (priority road) and class 17 (no entry). (b) Comparison of CORGI lower bounds and attack bounds for 100 random samples for each class. (c) Histogram of the ratio of attack upper bounds to CORGI lower bounds, with a median of 4.7 for class 12 and 5.0 for class 17.}
\end{figure}
\paragraph{Effectively choosing values of $k$ for interpretability maps} As people seek to produce accurate and descriptive interpretability maps, one important question is how $k$ should be chosen to faithfully represent the important regions of an input image. If $k$ is too small, then the top-$k$ CAM map may not fully capture the salient regions. If $k$ is too large, then we might capture noise pixels that were not pertinent to the model's prediction. Figure 3(a) shows a sample image and it's top-$k$ CAM map with increasing $k$ values.

One natural way to evaluate a specific value of $k$ is to consider $I(x)_{[k]}-I(x)_{[k+1]}$, which we call the \textit{CAM gap} for position $k$. Intuitively, if for some $k$, $I(x)_{[k]}$ and $I(x)_{[k+1]}$ are close, then only a small perturbation should be needed for $I(x)_{[k]}$ and $I(x)_{[k+1]}$ to swap places. Since the $k$th and $(k+1)$st largest pixel might not be well distinguished enough, that value of $k$ might be a poor choice. We would also expect that such a $k$ would be easily attacked, and more generally, that top-$k$ interpretability maps for values of $k$ with a larger CAM gap will require a larger perturbation to successfully attack. 

We varied $k$ from $1$ to $50$ in order to see how the CAM map gap was correlated to the performance of both our algorithm and the attack. Figure 3b demonstrates that as the CAM gap increases, the CORGI certificates and attack bounds both increase. This captures the intuition that the larger the CAM gap for a value $k$, the more robust the top-$k$ CAM interpretability map will be.

\begin{figure}[htbp]
  \begin{subfigure}[t]{0.15\textwidth}
    \centering
    \includegraphics[width=\linewidth]{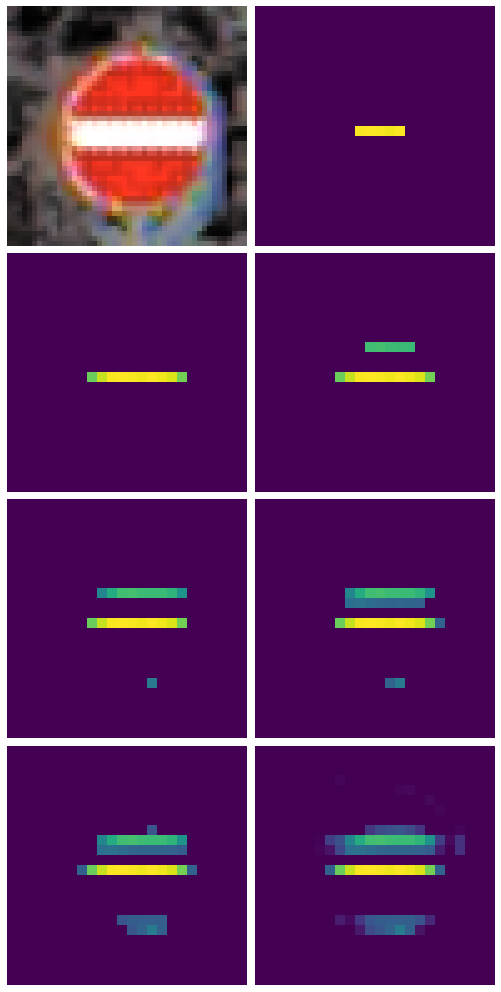}
    \caption{}
  \end{subfigure}
  \begin{subfigure}[t]{0.43\textwidth}
    \centering
    \includegraphics[width=\linewidth]{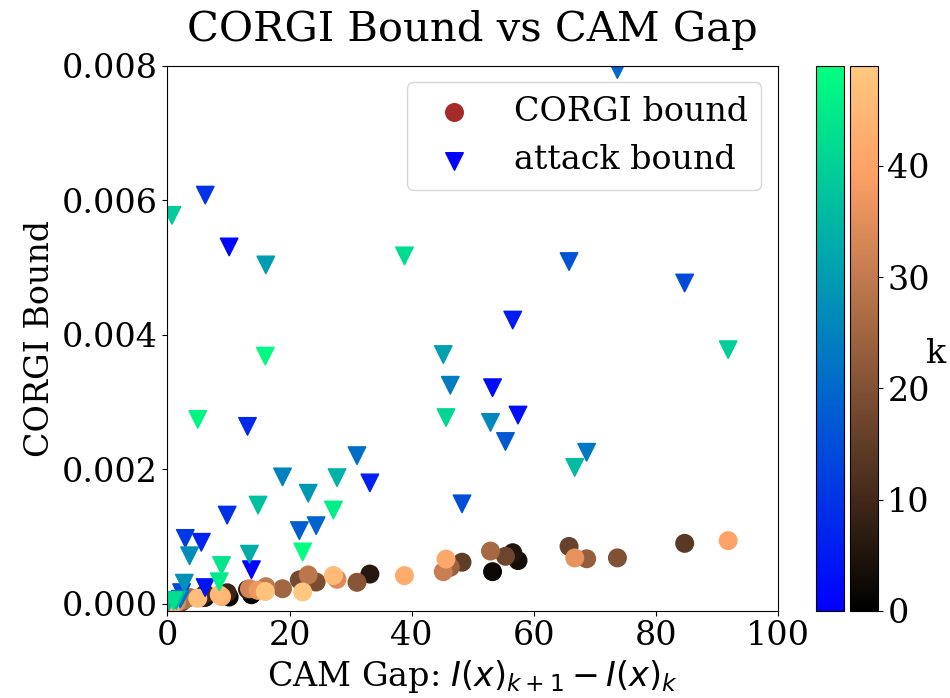}
    \caption{}
  \end{subfigure}
  \begin{subfigure}[t]{0.40\textwidth}
    \centering
    \includegraphics[width=\linewidth]{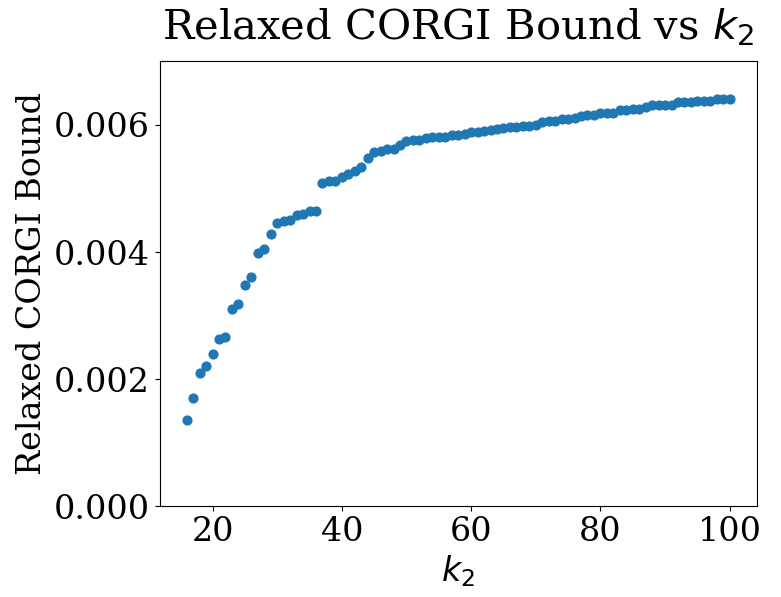}
    \caption{}
  \end{subfigure}
\centering
\caption{(a) The top-$k$ CAM map for an image with increasing values of $k$ from $5$ to $70$. (b) Both CORGI certificates and attack bounds are highly correlated with the CAM gap, meaning that choosing a $k$ with a large CAM gap leads to better interpretability robustness. (c) Relaxed CORGI bounds as a function of $k_2$, the maximum position of an original top-$k$ pixel in the perturbed image.}
\end{figure}
\paragraph{Relaxing the top-$k$ condition} We ran the modified CORGI algorithm given in Section 3.3 for a fixed image and fixed $k$. We varied the threshold $k_2$ to see how increasing $k_2$ would affect CORGI's bound. The results are shown in Figure 3c. For $k_2 \approx k$, increasing $k_2$ by just a small amount is able to strengthen our certificate by quite a bit. However, as $k_2$ becomes larger, the CORGI bounds begin to level off. We believe that this occurs because 

\section{Conclusions and Future Work}
We have presented CORGI, the first certifiable interpretability robustness for CAM. By using CNN-Cert to compute last layer convolutional output bounds, CORGI is able to efficiently provide a certificate $r \le r^*$. Through comparing our bounds with upper bounds generated by the iterative feature importance attack algorithm, CORGI provides bounds that are not far from certifiable upper bounds and acts as a solid baseline for future work in interpretability robustness. 

A limitation of CORGI is that it only performs well for small networks. Even for small networks, CORGI takes about $1$ minute per image. GTSRB images are significantly lower in resolution than images seen by autonomous vehicles, and networks used in the area are generally much larger, so CORGI does not yet scale to real-world applications. The main accuracy bottleneck of CORGI lies in the absence of fast and tight bounds for large convolutional neural networks. Therefore, improving both layer-wide network bounds and interpretability bounds for larger networks is a crucial direction to explore. 

We have not yet addressed certifiable robustness for other interpretability methods past CAM, such as DeepLIFT, Integrated Gradients, and GradCAM. Since CAM is a bit restrictive in the architectures that it is able to handle and requires a global average pooling layer following convolutional layers, developing bounds for these other interpretability algorithms will expand the class of models for which we can develop certificates.

\section{Acknowledgement}
Alex Gu, Tsui-Wei Weng and Luca Daniel are partially supported by MIT-IBM Watson AI Lab.

\newpage
\renewcommand{\bibsection}{\section{\bibname}}
\typeout{}
\bibliography{mybib.bib}
\bibliographystyle{plainnat}

\newpage
\section{Appendix}

\subsection{Model and Experimental Details}

The architecture of our neural network is shown below. Note that the CAM map is a weighted average of the $16$ filters in the last convolutional layer, so has shape $24 \times 24$. The test accuracy of our network is around 92\%. Ghorbani's attack algorithm was run with $P=300$ steps and $\alpha = \epsilon/10$. Increasing $P$ did not have a significant impact on the effectiveness of the attack. 

\begin{table}[htbp]
\centering
\begin{tabular}{|c|lll} 
\cline{1-1}
\textbf{Input $(48 \times 48 \times 3)$} & & & \\
\cline{1-1}
Conv with 8 filters $(48 \times 48 \times 8)$  &  &  &   \\ 
\cline{1-1}
ReLU                              &  &  &   \\ 
\cline{1-1}
Conv with 8 filters $(48 \times 48 \times 8)$  &  &  &   \\ 
\cline{1-1}
ReLU                              &  &  &   \\ 
\cline{1-1}
Max Pool $(24 \times 24 \times 8)$                       &  &  &   \\ 
\cline{1-1}
Dropout $(0.2)$                     &  &  &   \\ 
\cline{1-1}
Conv with 16 filters $(24 \times 24 \times 16)$ &  &  &   \\ 
\cline{1-1}
ReLU                              &  &  &   \\ 
\cline{1-1}
Conv with 16 filters $(24 \times 24 \times 16)$ &  &  &   \\ 
\cline{1-1}
ReLU                              &  &  &   \\ 
\cline{1-1}
Global Average Pooling   $(16)$         &  &  &   \\ 
\cline{1-1}
Dense               $(43)$              &  &  &   \\
\cline{1-1}
\end{tabular}
\end{table}
\subsection{Relaxed CAM Certificate}
In this section, we give the proof of Theorems 3.3 in our main work.

\paragraph{Theorem 3.3 (Relaxed CAM Interpretability Certificate)} \textit{Let $x \in \R^{s \times s}$ be a fixed input image, $\delta \in \R_{\ge 0}$ be a fixed maximum perturbation size, $k, k_2 \in \mathbb{Z}_{\ge 1}$ such that $k_2 \ge k$, and $L(\delta), R(\delta)$ be CAM map bounds as computed in Eq. 1. Then, for all $x'$ such that $\|x-x'\|_{\infty} \le \delta$, if \begin{equation}\min_{k' \in 1, 2, \cdots, k} L(\delta)_{\sigma_{k'}} \ge U(\delta)_{\{k_2-k+1\}}\end{equation} then under all perturbations of size at most $\delta$, the top-$k$ pixel positions of the original CAM map $I(x)$ are in the top $k_2$ pixel positions of the CAM map $I(x')$.}

\paragraph{Proof} The proof is very similar to that of Theorem 3.2. We have the following chain of inequalities:
\begin{equation*}\min_{k' \in 1, 2, \cdots, k} I(x')_{\sigma_{k'}} \stackrel{(1)}{\ge} \min_{k' \in 1, 2, \cdots, k} L(\delta)_{\sigma_{k'}}\stackrel{(2)}{\ge} U(\delta)_{\{k_2-k+1\}} \stackrel{(3)}{=} U(\delta)_{[k_2+1]} \stackrel{(4)}{\ge} I(x')_{[k_2+1]}\end{equation*}

$(1)$ comes from Eq. 3, $(2)$ comes from the hypothesis, and (4) is an implication of Eq. 1. To see why $(3)$ holds, note that the hypothesis guarantees that $U(\delta)_{\sigma_1}, \cdots, U(\delta)_{\sigma_k} \ge U(\delta)_{\{k_2-k+1\}}$, which means that the $(k_2+1)$st largest pixel value of $U(\delta)$ must be equal to the $(k_2-k+1)$st largest value of $U(\delta)$ without considering the $\sigma_i$'s. 

\subsection{Lipschitz Constant CAM Certificate}
In this section, we give a different theoretical bound directly based on the Lipschitz constant of a neural network.

First, recall the definition of a locally Lipschitz function: \paragraph{Definition 3.4.1:} A function $f:\R^d \to \R$ is \textit{L-locally Lipschitz} in a region $S \subseteq \R^d$ if for all $x, y \in S$, we have $|f(x)-f(y)|_1 \le L\|x-y\|_{\infty}$. The smallest $L$ for which this inequality is true is called the \textit{local Lipschitz constant} of $f$ in a region $S$. 

We can then guarantee the following certificate, the proof of which will be deferred to the Appendix.
\paragraph{Theorem 8.3 (Lipschitz-Based CAM Interpretability Certificate)}\textit{
Let $x \in \R^{s \times s}$ be a fixed input image, $I( \cdot ) : \R^{n \times n} \to \R^{s \times s}$ be the CAM interpretability mapping function, $k \in \mathbb{Z}_{\ge 1}$, and $\{\sigma_i\}_{1 \le i \le k}$ denote the top-$k$ pixels of the input image $I(x)$. For $1 \le i \le k$, $k+1 \le p \le s^2$, define $g_p^{(i)}(x) \triangleq I(x)_{\sigma_i} - I(x)_p$, and let $L_p^{(i)}(\delta)$ be the local Lipschitz constant of the function $g_p^{(i)}$ in some region $S = \{x'|\delta \ge \|x-x'\|_{\infty}\}$ for some $\delta \in \R$. Finally, let $$r = \min_{k+1 \le p \le s^2}\left( \min_{1 \le i \le k} \frac{g_p^{(i)}(x)}{L_p^{(i)}(\delta)}\right)$$ Then, for all $x$ such that $$\|x-x'\|_{\infty} \le r$$ the set of top-$k$ pixels of $I(x)$ will be identical to that of $I(x')$ as long as $\delta$ is chosen such that $\delta \ge r$. In other words, $r$ is a certified radius for top-$k$ interpretability.}

\paragraph{Proof} Note that in order to have the top-$k$ pixel positions of $I(x)$ be identical to that of $I(x')$, we must have, for all $1 \le i \le k$ and $k+1 \le p \le s^2$, $$g_p^{(i)}(x') = I(x')_{\sigma_i} - I(x')_{p} \ge 0$$ This is essentially a restatement of Eq. 4. By the definition of the Lipschitz constant, for any $\epsilon$ such that $\|\epsilon\|_{\infty} < \delta$, we have $$|g_p^{(i)}(x+\epsilon) - g_p^{(i)}(x)|_1 \le L_p^{(i)}(\delta) \| \epsilon\|_{\infty}$$
$$\Rightarrow g_p^{(i)}(x) - L_p^{(i)}(\delta) \|\epsilon\|_{\infty} \le g_p^{(i)}(x+\epsilon)$$

Observe that if $\|\epsilon\|_{\infty} \le \frac{g_p^{(i)}(x)}{L_p^{(i)}(\delta)}$, then $g_p^{(i)}(x+\epsilon) \ge 0$. Therefore, if $$\|\epsilon\|_{\infty} \le \min_{k+1 \le p \le s^2} \left(\min_{1 \le i \le k} \frac{g_p^{(i)}(x)}{L_p^{(i)}(\delta)} \right)$$ then $g_p^{(i)}(x+\epsilon) \ge 0$ for all $1 \le i \le k, k+1 \le p \le s^2$, which is precisely the condition that the top-$k$ pixels of $I(x)$ are identical to the top-$k$ pixels of $I(x+\epsilon)$ under $\|\epsilon\|_{\infty} \le r$.

In its essence, CORGI provides certificates by relying on CNN-Cert to derive bounds on the interpretability maps and then ensuring that the lower bound of $I(x')_{\sigma_i}$ is larger than the upper bound of the other pixels. For larger networks like AlexNet and ResNets, however, CNN-Cert's bounds are often from known attack bounds ~\citep{boopathy2019cnn}, and are suspected to be loose. Therefore, Theorem 8.3 provides a more straightforward approach by directly operating on the Lipschitz constant of the function $g_p^{(i)}$. Therefore, for tight estimates of the Lipschitz constant, we expect this bound to outperform CORGI. However, Lipschitz constant estimation is currently still a nascent and young area of research~\citep{jordan2020exactly}~\citep{zhang2019recurjac}~\citep{fazlyab2019efficient}. Therefore, as current methods for estimating Lipschitz constants are either very loose or run very slowly, we will not give an empirical analysis of this approach.

\subsection{Additional Image Examples}

We show two more examples of images, CORGI bounds, and CAM maps similar to Figure 1.

\begin{figure}[h]
\centering
\includegraphics[width=0.9\linewidth]{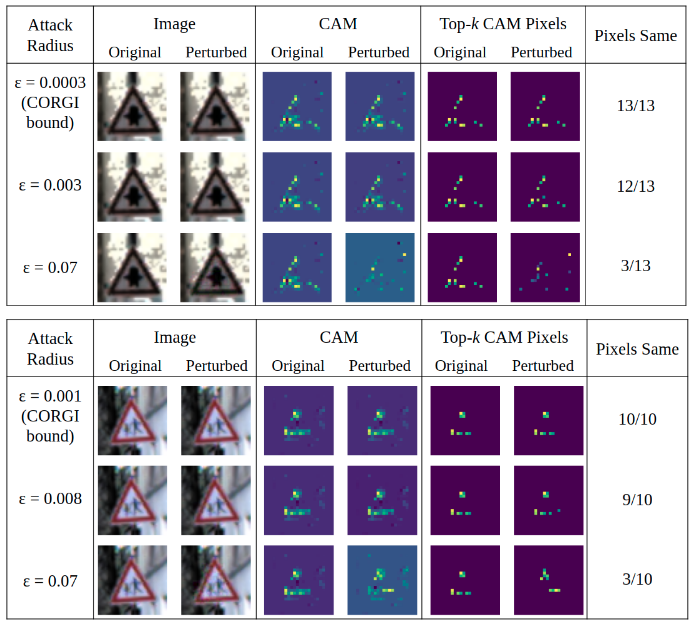}

\caption{GTSRB images under 3 different attack radii. Shown are the original and perturbed images, CAM maps, and the top-$k$ CAM pixel positions.}
\end{figure}

\subsection{Additional CORGI Examples}
We show a comparison between attack and CORGI bounds for 5 additional classes, similar to that of Figure 2. For each of the 5 classes, 100 test images were random chosen and evaluated. The distribution of ratios is similar for all of the classes, suggesting that CORGI performs fairly uniformly throughout the entire testing distribution.

\begin{figure}[H]
\centering
\includegraphics[width=\linewidth]{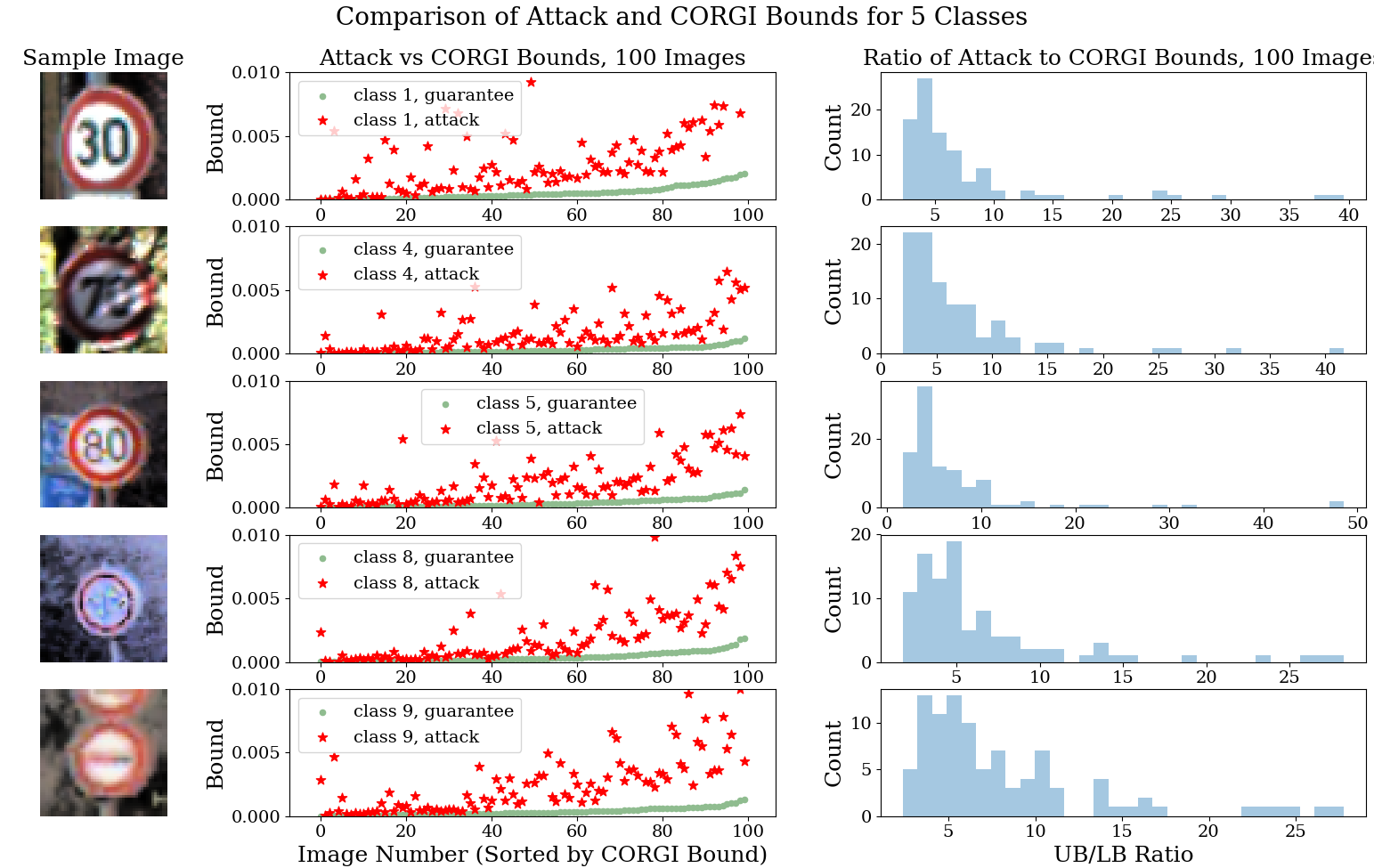}

\caption{Comparison of CORGI vs attack bounds for 5 additional classes.}
\end{figure}
\end{document}